\documentclass[letterpaper, 10pt, conference]{ieeeconf}      
 
\IEEEoverridecommandlockouts                              

\usepackage{algorithm}
\usepackage{multicol}
\usepackage{multirow}
\usepackage{algpseudocode}
\usepackage{color}
\usepackage{xcolor}
\usepackage{graphicx}
\usepackage{bm}
\usepackage{amsmath}
\usepackage{amssymb}
\usepackage{tabularx}
\usepackage{subcaption}
\usepackage{url}
\usepackage{cite}

\newcommand{\minus}{\scalebox{0.75}[1.0]{$-$}}
\DeclareMathOperator*{\argmin}{\arg\!\min}

\usepackage{hyperref}
\hypersetup{
    colorlinks,
    linkcolor={blue!80!black},
    citecolor={blue!80!black},
    urlcolor={blue!100!black}
}

\title{\LARGE  \bf
Towards Optimally Decentralized Multi-Robot Collision Avoidance via Deep Reinforcement Learning} 

\author{Pinxin Long$^{1}$*, Tingxiang Fan$^{1}$*, Xinyi Liao$^{1}$, Wenxi Liu$^{2}$, Hao Zhang$^{1}$ and Jia Pan$^{3}$%
\thanks{* denotes equal contribution.}
\thanks{$^{1}$Pinxin Long, Tingxiang Fan, Xinyi Liao and Hao Zhang are with Dorabot Inc., Shenzhen, China {\tt\small pinxinlong@gmail.com}}%
\thanks{$^{2}$Wenxi Liu is with the Department of Computer Science, Fuzhou University, Fuzhou, China {\tt\small wenxi.liu@hotmail.com}}%
\thanks{$^{3}$Jia Pan is with the Department of Mechanical and Biomedical Engineering, City University of Hong Kong, Hong Kong, China {\tt\small jiapan@cityu.edu.hk}}%
\thanks{This paper is partially supported by HKSAR General  Research  Fund (GRF) CityU 21203216, and NSFC/RGC Joint Research Scheme (CityU103/16-NSFC61631166002)}
}

\begin{document}
\maketitle

\begin{abstract}
Developing a safe and efficient collision avoidance policy for multiple robots is challenging in the decentralized scenarios where each robot generates its paths without observing other robots' states and intents. While other distributed multi-robot collision avoidance systems exist, they often require extracting agent-level features to plan a local collision-free action, which can be computationally prohibitive and not robust. More importantly, in practice the performance of these methods are much lower than their centralized counterparts. 

We present a decentralized sensor-level collision avoidance policy for multi-robot systems, which directly maps raw sensor measurements to an agent's steering commands in terms of movement velocity. As a first step toward reducing the performance gap between decentralized and centralized methods, we present a multi-scenario multi-stage training framework to learn an optimal policy. The policy is trained over a large number of robots on rich, complex environments simultaneously using a policy gradient based reinforcement learning algorithm. 
We validate the learned sensor-level collision avoidance policy in a variety of simulated scenarios with thorough performance evaluations and show that the final learned policy is able to find time efficient, collision-free paths for a large-scale robot system. We also demonstrate that the learned policy can be well generalized to new scenarios that do not appear in the entire training period, including navigating a heterogeneous group of robots and a large-scale scenario with 100 robots. Videos are available at \url{https://sites.google.com/view/drlmaca}.
\end{abstract}



\section{Introduction}
\label{sec:intro}


Multi-robot navigation has recently gained much interest in robotics and artificial intelligence, and has many real-world applications including multi-robot search and rescue, navigation through human crowds, and autonomous warehouse. One of the major challenges for multi-robot navigation is to develop a safe and robust collision avoidance policy for each robot navigating from its starting position to its desired goal. 

Some of prior works, known as \textit{centralized methods}, assume that the comprehensive knowledge about all agents' intents (e.g. initial states and goals) and their workspace (e.g. a 2D grid map) is given for a central server to control the action of agents. These methods can generate the collision avoidance action by planning optimal paths for all robots simultaneously. However, these centralized methods are difficult to scale to large systems with many robots and their performance can be low when frequent task/goal reassignment is necessary. Besides, in practice, they heavily rely on the reliable communication network between robots and the central server. Therefore, once the central server and/or the communication network fails, the multi-robot systems will break down. Furthermore, these centralized methods are inapplicable when multiple robots are deployed in an unknown and unstructured environments.  

\begin{figure}[t] 
\centering
\includegraphics[width=1.0\linewidth]{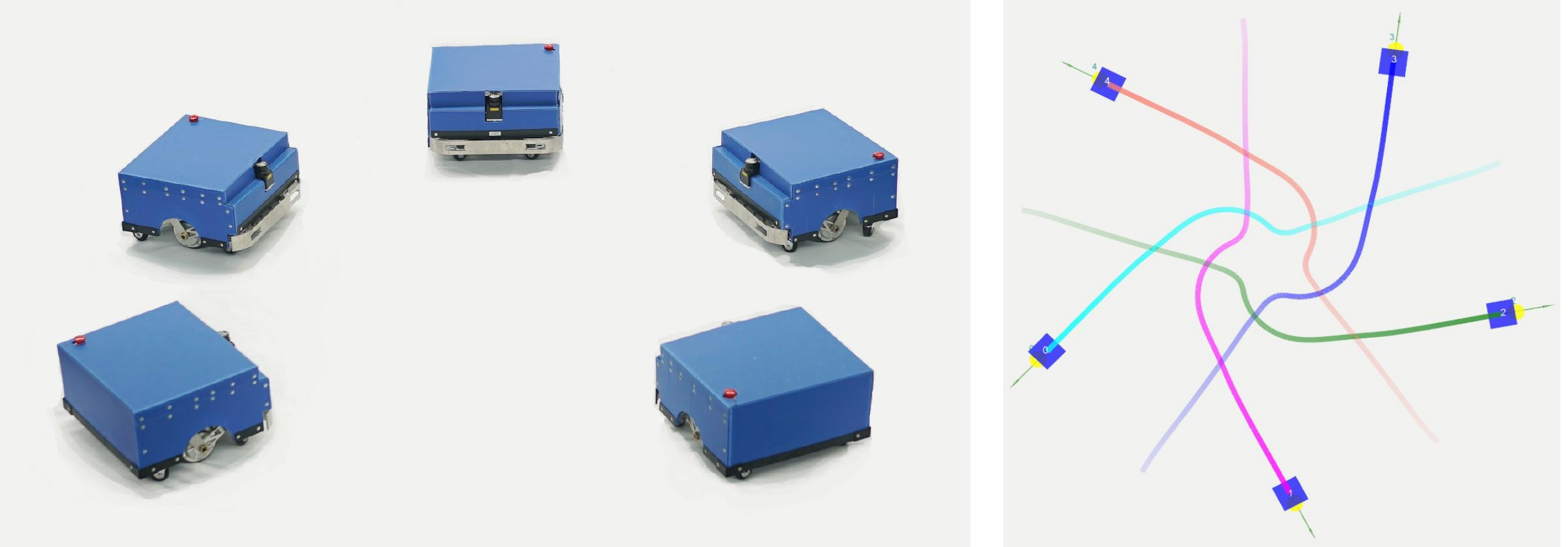} 
\caption{Robot trajectories in the \emph{Circle} scenario using our learned policy. Note that the robots are square-shaped. It shows the good generalization capability of the learned policy since we directly test the policy trained with disc-shaped robots on this scenarios.}
\label{fig:first}
\vspace*{-0.3in}
\end{figure}

Compared with centralized methods, some existing works propose \textit{agent-level decentralized collision avoidance} policies, where each agent independently makes decision taking into account the observable states (e.g. shapes, velocities and positions) of other agents as inputs.
Most agent-level policies are based on the velocity obstacle (VO)~\cite{ Berg:ORCA:2011, snape2011hybrid, hennes2012multi, claes2012collision, bareiss2015generalized}, and they can compute local collision-free action efficiently for multiple agents in cluttered workspaces.  
However, several limitations greatly restrict their applications. 
First, the simulation based works~\cite{van2008reciprocal,Berg:ORCA:2011} assume that each agent has perfect sensing about the surrounding environment which does not hold in real world scenarios due to omnipresent sensing uncertainty. 
To moderate the limitation of perfect sensing, previous approaches use a global positioning system to track the positions and velocities of all robots~\cite{snape2011hybrid, bareiss2015generalized} or design an inter-agent communication protocol for sharing position and velocity information among nearby agents~\cite{claes2012collision, hennes2012multi, godoy2016implicit}. However, these approaches introduce external tools or communication protocols into the multi-robot systems, which may not be adequately robust. Second, VO based policies have many tunable parameters that are sensitive to scenario settings and thus the parameters must be carefully set offline to achieve satisfying performance. Finally, the performance of previous decentralized methods in terms of navigation speed and navigation time is significantly lower than their centralized counterparts. 

Inspired by VO based approaches, Chen et al.~\cite{chen2017decentralized} train an agent-level collision avoidance policy using deep reinforcement learning, which learns a two-agent value function that explicitly maps an agent's own state and its neighbors' states to collision-free action, whereas it still demands the perfect sensing. In their later work~\cite{chen2017socially}, multiple sensors are deployed to perform tasks of segmentation, recognition, and tracking in order to estimate the states of nearby agents and moving obstacles. However, this complex pipeline not only requires expensive online computation but makes the whole system less robust to the perception uncertainty.  

In this paper, we focus on 
\textit{sensor-level decentralized collision avoidance} policies that directly map the raw sensor data to desired, collision-free steering commands. Compared with agent-level policies, the perfect sensing for the neighboring agents and obstacles, and offline parameter-tuning for different scenarios are not required. 
Sensor-level collision avoidance policies are often modeled by deep neural networks (DNNs)~\cite{long2017deep,pfeiffer2017perception} and trained using supervised learning on a large dataset.  
However, there are several limitations for learning policies under supervision. First, it requires a large amount of training data that should cover different kinds of interaction situations for multiple robots. Second, the expert trajectories in datasets are not guaranteed to be optimal in the interaction scenarios, which makes training difficult to converge to a robust solution.  
Third, it is difficult to hand-design a proper loss function for training robust collision avoidance policies. To overcome these drawbacks, we propose a multi-scenario multi-stage deep reinforcement learning framework to learn the optimal collision avoidance policy using the policy gradient method. 

\noindent \textbf{Main results:} 
In this paper, we address the collision avoidance of multiple robots in a fully decentralized framework, in which the input data is only collected from onboard sensors.
To learn the optimal collision avoidance policy, we propose a novel multi-scenario multi-stage training framework which exploits a robust policy gradient based reinforcement learning algorithm trained in a large-scale robot system in a set of complex environments. 
We demonstrate that the collision avoidance policy learned from the proposed method is able to find time efficient, collision-free paths for a large-scale nonholonomic robot system, and it can be well generalized to unseen scenarios. Its performance is also much better than previous decentralized methods, and can serve as a first step toward reducing the gap between centralized and decentralized navigation policies.

\begin{figure}
\centering
\includegraphics[width=0.6\linewidth]{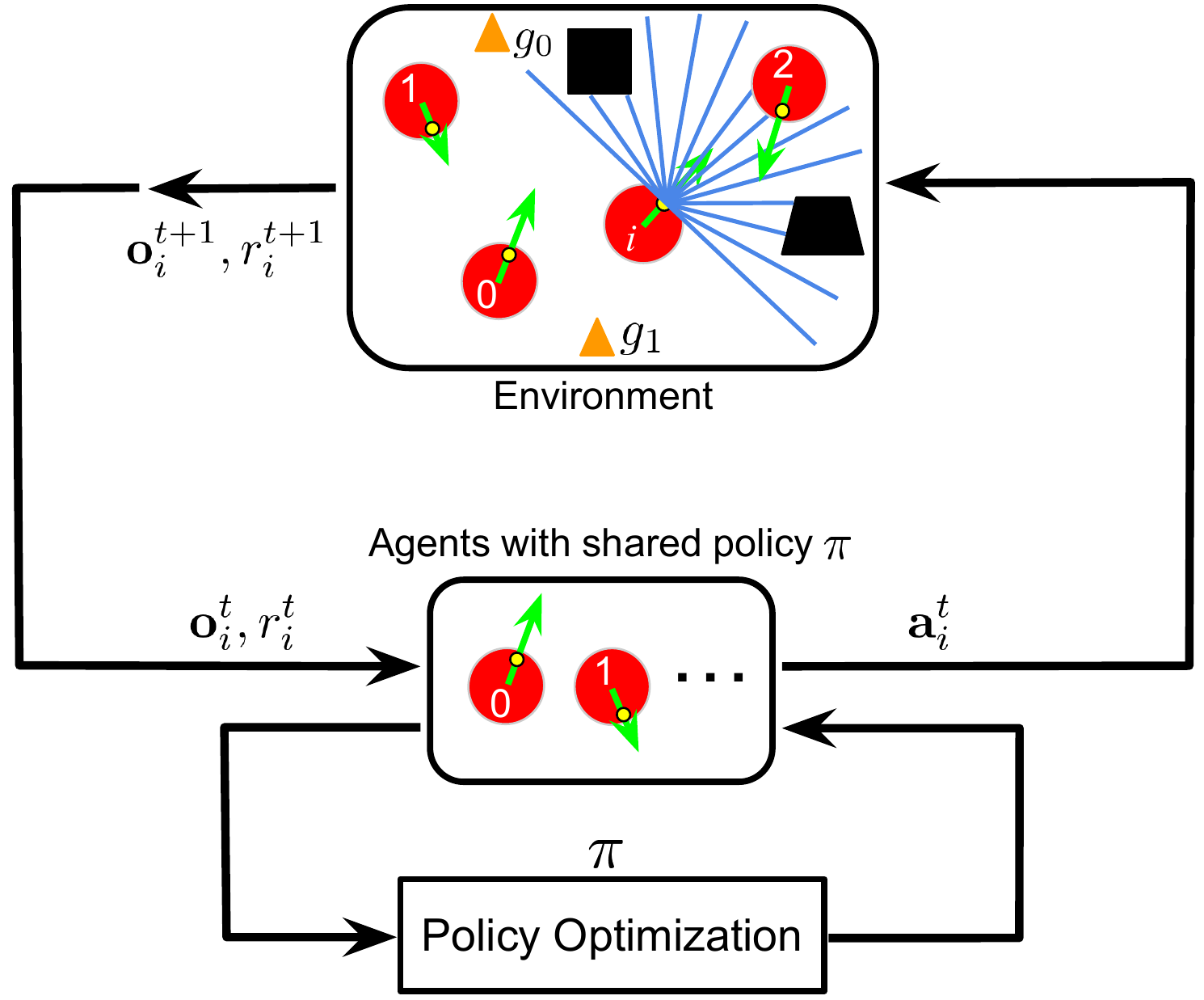} 
\caption{An overview of our approach. At each timestep $t$, each robot receives its observation $\mathbf{o}^t_i$ and reward $r^t_i$ from the environment, and generates an action $a_i^t$ following the policy $\pi$. The policy $\pi$ is shared across all robots and updated by a policy gradient based reinforcement learning algorithm.}
\label{fig:overview}
\vspace*{-0.3in}
\end{figure}
\section{Related Work}
\label{sec:related}
Learning-based collision avoidance techniques have been extensively studied for one robot avoiding static obstacles. Many approaches adopt the supervised learning paradigm to train a collision avoidance policy by imitating a dataset of sensor input and motion commands.
Muller et al.~\cite{muller2006off} trained a vision-based static obstacle avoidance system in supervised mode for a mobile robot by training a 6-layer convolutional network which maps raw input images to steering angles. 
Zhang et al.~\cite{zhang2016deep} exploited a successor-feature-based deep reinforcement learning algorithm to transfer depth information learned in previously mastered navigation tasks to new problem instances.
Sergeant et al.~\cite{sergeant2015multimodal} proposed a mobile robot control system based on multimodal deep autoencoders. 
Ross et al.~\cite{ross2013learning} trained a discrete controller for a small quadrotor helicopter with imitation learning techniques. The quadrotor was able to successfully avoid collisions with static obstacles in the environment using only a single cheap camera. Yet only discrete movement (left/right) has to be learned and also the robot only trained within static obstacles. 
Note that the aforementioned approaches only take into account the static obstacles and require a human driver to collect training data in a wide variety of environments. 
Another data-driven end-to-end motion planner is presented by Pfeiffer et al.~\cite{pfeiffer2017perception}. They trained a model maps laser range findings and target positions to motion commands using expert demonstrations generated by the ROS navigation package. This model can navigate the robot through a previously unseen environment and successfully react to sudden changes. Nonetheless, similar to the other supervised learning methods, the performance of the learned policy is seriously constrained by the quality of the labeled training sets. 
To overcome this limitation, Tai et al.~\cite{tai2017virtual} proposed a mapless motion planner trained through a deep reinforcement learning method. Kahn et al.~\cite{kahn2017uncertainty} presented an uncertainty-aware model-based reinforcement learning algorithm to estimate the probability of collision in a priori unknown environment. However, the test environments are relative simple and structured, and the learned planner is hard to generalize to the scenarios with dynamic obstacles and other proactive agents.


Regarding \textit{multi-agent} collision avoidance, the Optimal Reciprocal Collision Avoidance (ORCA) framework~\cite{Berg:ORCA:2011} has been popular in crowd simulation and multi-agent systems. ORCA provides a sufficient condition for multiple robots to avoid collisions with each other in a short time horizon, and can easily be scaled to handle large systems with many robots. ORCA and its extensions~\cite{snape2011hybrid,bareiss2015generalized} used heuristics or first principles to construct a complex model for the collision avoidance policy, which has many parameters that are tedious and difficult to be tuned properly. Besides, these methods are sensitive to the uncertainties ubiquitous in the real-world scenarios since they assume each robot to have perfect sensing about the surrounding agents' positions, velocities and shapes. To alleviate the requirement of perfect sensing, communication protocols are introduce by~\cite{hennes2012multi,claes2012collision,godoy2016implicit} to share the state information including agents' positions and velocities among the group.
Moreover, the original formulation of ORCA is based on holonomic robots which is less common than nonholonomic robots in the real world. To deploy ORCA on the most common differential drive robots, several methods have been proposed to deal with the difficulty of non-holonomic robot kinematics. ORCA-DD~\cite{snape2010smooth} enlarges the robot to twice the radius of the original size to ensure collision free and smooth paths for robots under differential constraints. However, this enlarged virtual size of the robot can result in problems in narrow passages or unstructured environments. 
NH-ORCA~\cite{alonso2013optimal} makes a differential drive robot tracking a holonomic speed vector with a certain tracking error $\varepsilon$. It is preferred over ORCA-DD, since the virtual increase of the robots' radii is only by a size of $\varepsilon$ instead of doubling the radii. 

In this paper, we focus on learning a collision-avoidance policy which can make multiple nonholonomic mobile robots navigate to their goal positions without collisions in rich and complex environments.

\section{Problem Formulation}
\label{sec:prob}
The multi-robot collision avoidance problem is defined primarily in the context of a nonholonomic differential drive robot moving on the Euclidean plane with obstacles and other decision-making robots. 
During training, all of $N$ robots are modeled as discs with the same radius $R$, i.e., all robots are homogeneous. 

At each timestep $t$, the $i$-th robot $(1 \leq i \leq N)$ has access to an observation $\mathbf o_{i}^t$ and computes a collision-free steering command $\mathbf a_{i}^t$ that drives it to approach the goal $\mathbf g_{i}$ from the current position $\mathbf p_{i}^t$.  The observation $\mathbf o_{i}^t$ drawn from a probability distribution w.r.t. the underlying system state $\mathbf s_{i}^t$, $\mathbf o_{i}^t \sim \mathcal{O}(\mathbf s_{i}^t)$, only provides partial information about the state $\mathbf s_{i}^t$ since the $i$-th robot has no explicit knowledge about other robots' states and intents. Instead of the perfect sensing assumption applied in prior methods (e.g.~\cite{chen2017decentralized,chen2017socially,claes2012collision,hennes2012multi,van2008reciprocal,Berg:ORCA:2011}), our formulation based on partial observation makes our approach more applicable and  robust in real world applications.
The observation vector of each robot can be divided into three parts: $\mathbf o^t = [\mathbf{o}_{z}^t, \mathbf{o}_{g}^t, \mathbf{o}_{v}^t]$ (here we ignore the robot ID $i$ for legibility), where $\mathbf{o}_{z}^t$ denotes the raw 2D laser measurements about its surrounding environment, $\mathbf{o}_{g}^t$ stands for its relative goal position (i.e. the coordinates of the goal in the robot's local polar coordinate frame), and $\mathbf{o}_{v}^t$ refers to its current velocity. 
Given the partial observation $\mathbf o^t$, each robot \textit{independently} computes an action or a steering command, $\mathbf{a}^t$, sampled from a stochastic policy $\pi$ shared by all robots: 
\begin{equation}
\label{eq:stpolicy}
  \mathbf{a}^t \sim \pi_{\theta}(\mathbf{a}^t \mid \mathbf{o}^t),
\end{equation}
where ${\theta}$ denotes the policy parameters. 
The computed action $\mathbf{a}^t$ is actually a velocity $\mathbf{v}^t$ that guides the robot approaching its goal while avoiding collisions with other robots and obstacles $\mathbf{B}_k$ $(0 \leq k \leq M)$ within the time horizon $\Delta t$ until the next observation $\mathbf o^{t+1}$ is received. 

Hence, the multi-robot collision avoidance problem can be formulated as a partially observable sequential decision making problem. The sequential decisions consisting of observations and actions (velocities) $(\mathbf{o}_i^t, \mathbf{v}_i^t)_{t=0:t^g_i}$ made by the robot $i$ can be considered as a trajectory $l_{i}$ from its start position $\mathbf{p}_i^{t=0}$ to its desired goal $\mathbf{p}_i^{t=t_i^g} \equiv \mathbf{g}_i$, where $t_i^g$ is the traveled time. To wrap up the above formulation, we define $\mathbb{L}$ as the set of trajectories for all robots, which are subject to the robot's kinematic (e.g. non-holonomic) constraints, i.e.:  
\begin{equation}
\begin{aligned}
\mathbb{L} ={} & \{ l_i, i = 1,..., N \mid \\
        & \mathbf{v}_i^t \sim \pi_{\theta}(\mathbf{a}_i^t \mid \mathbf{o}_i^t), \\
        & \mathbf{p}_i^t = \mathbf{p}_i^{t-1} + \Delta t \cdot \mathbf{v}_i^{t},  \\
        & \forall j \in [1, N], j\neq i: \| \mathbf{p}_i^t - \mathbf{p}_j^t \| > 2R \\
        & \land \forall k \in [1, M]: \| \mathbf{p}_i^t - \mathbf{B}_k \| > R   \\
        & \land \| \mathbf{v}_i^t \| \leq v_i^{\mathrm{max}} \}.
\end{aligned}
\end{equation}

To find an optimal policy shared by all robots, we adopt an objective by minimizing the expectation of the mean arrival time of all robots in the same scenario, which is defined as:
\begin{align}
\argmin_{\pi_{\theta}}~~\mathbb{E}[\frac{1}{N}\sum_{i=1}^N t_i^g|{\pi}_{\theta}],
\end{align}
where $t_i^g$ is the travel time of the trajectory $l_i$ in $\mathbb{L}$ controlled by the shared policy $\pi_{\theta}$.

The average arrival time will also be used as an important metric to evaluate the learned policy in Section~\ref{sec:exp}.
We solve this optimization problem through a policy gradient based reinforcement learning method, which bounds the policy parameter updates to a trust region to ensure stability.

\section{Approach}
\label{sec:approach}
We begin this section by introducing the key ingredients of our reinforcement learning framework. Next, we describe the details about the architecture of the collision avoidance policy in terms of a deep neural network. Finally, we elaborate the training protocols used to optimize the policy.

\subsection{Reinforcement Learning Setup}
\label{sec:setup}
The partially observable sequential decision making problem defined in Section~\ref{sec:prob} can be formulated as a Partially Observable Markov Decision Process (POMDP) solved with reinforcement learning. Formally, a POMDP can be described as a 6-tuple $(\mathcal{S}, \mathcal{A}, \mathcal{P}, \mathcal{R}, \Omega, \mathcal{O})$, where $\mathcal{S}$ is the state space, $\mathcal{A}$ is the action space, $\mathcal{P}$ is the state-transition model, $\mathcal{R}$ is the reward function, $\Omega$ is the observation space ($\mathbf o \in \Omega$) and $\mathcal{O}$ is the observation probability distribution given the system state ($\mathbf o \sim \mathcal{O}(\mathbf s)$). In our formulation, each robot only has access to the observation sampled from the underlying system states. Furthermore, since each robot plans its motions in a fully decentralized manner, a multi-robot state-transition model $\mathcal{P}$ determined by the robots' kinematics and dynamics is not needed. Below we describe the details of the observation space, the action space, and the reward function. 

\subsubsection{\textbf{Observation space}} 
As mentioned in Section~\ref{sec:prob}, the observation $\mathbf{o}^t$ consists of the readings of the 2D laser range finder $\mathbf{o}_z^t$, the relative goal position $\mathbf{o}_g^t$ and robot's current velocity $\mathbf{o}_v^t$. Specifically, $\mathbf{o}_z^t$ includes the measurements of the last three consecutive frames from a 180-degree laser scanner which has a maximum range of 4 meters and provides 512 distance values per scanning (i.e. $\mathbf{o}_z^t \in \mathbb{R}^{3 \times 512}$). In practice, the scanner is mounted on the forepart of the robot instead of the center (see the left image in Figure~\ref{fig:first}) to obtain a large unoccluded view. The relative goal position $\mathbf{o}_g^t$ is a 2D vector representing the goal in polar coordinate (distance and angle) with respect to the robot's current position. The observed velocity $\mathbf{o}_v^t$ includes the current translational and rotational velocity of the differential-driven robot. The observations are normalized by subtracting the mean and dividing by the standard deviation using the statistics aggregated over the course of the entire training.  
\subsubsection{\textbf{Action space}} 
The action space is a set of permissible velocities in continuous space. The action of differential robot includes the translational and rotational velocity, i.e. $\mathbf{a}^{t} = [v^{t}, w^{t}]$. In this work, considering the real robot's kinematics and the real world applications, we set the range of the translational velocity $v \in (0.0, 1.0)$ and the rotational velocity in $w \in (-1.0, 1.0)$. Note that backward moving (i.e. $v < 0.0$) is not allowed since the laser range finder can not cover the back area of the robot. 
\subsubsection{\textbf{Reward design}} 
Our objective is to avoid collisions during navigation and minimize the mean arrival time of all robots. A reward function is designed to guide a team of robots to achieve this objective: 
\begin{equation}
r_i^t = (^gr)_i^t + (^cr)_i^t + (^wr)_i^t. 
\end{equation}
The reward $r$ received by robot $i$ at timestep $t$ is a sum of three terms, $^gr$, $^cr$, and $^wr$. In particular, the robot is awarded by $(^gr)_i^t$ for reaching its goal:
\begin{equation} 
(^gr)_i^t = 
\begin{cases}
  r_{arrival}  & \ \text{if } \| \mathbf{p}_i^t - \mathbf{g}_i \| < 0.1 \\
  \omega_g(\|\mathbf{p}_i^{t-1}- \mathbf{g}_i \| - \|\mathbf{p}_i^{t}- \mathbf{g}_i \|) & \ \text{otherwise}. \\
\end{cases}
\end{equation}
When the robot collides with other robots or obstacles in the environment, it is penalized by $(^cr)_i^t$: 
\begin{equation}
(^cr)_i^t =  
\begin{cases}
r_{collision} & \quad \text{if } \| \mathbf{p}_i^t - \mathbf{p}_j^t \| < 2R \\ 
              & \quad \text{or } \|\mathbf{p}_i^t - \mathbf{B}_k\| < R \\
0 & \quad \text{otherwise}.  \\
\end{cases}
\end{equation}
To encourage the robot to move smoothly, a small penalty $(^wr)_i^t$ is introduced to punish the large rotational velocities: 
\begin{equation}  
(^wr)_i^t = \omega_{w}|w_i^t|  \quad \quad \text{if }  |w_i^t| > 0.7 . 
\end{equation}
We set $r_{arrival}=15$, $\omega_g=2.5$, $r_{collision}=\minus 15$ and $\omega_w=\minus 0.1$ in the training procedure.

\subsection{Network architecture}
\label{sec:model}

Given the input (observation $\mathbf{o}_i^t$) and the output (action $\mathbf{v}_i^t$), now we elaborate the policy network mapping $\mathbf{o}_i^t$ to $\mathbf{v}_i^t$. 

\begin{figure}
\centering 
\includegraphics[width=0.8\linewidth]{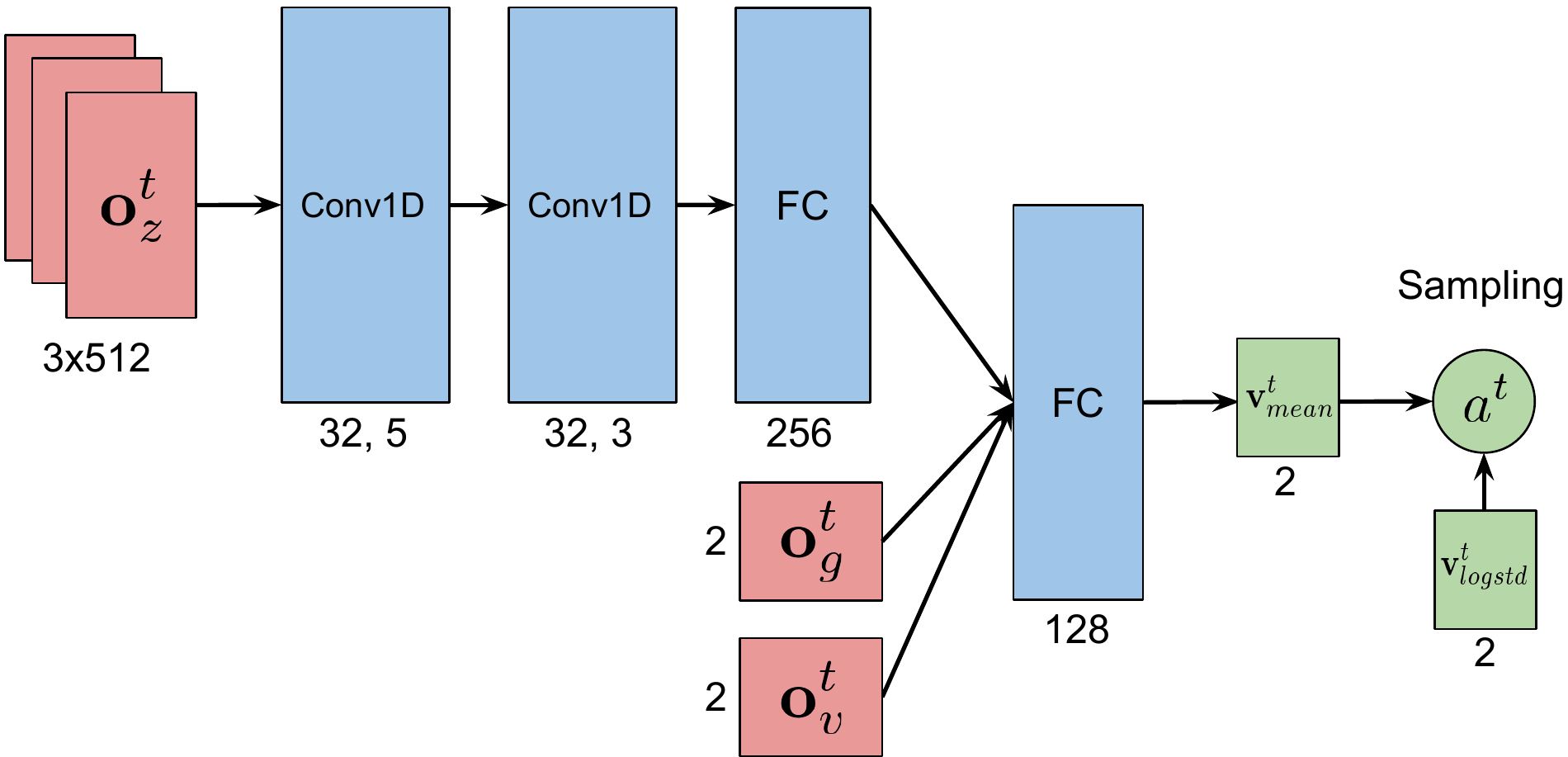} 
\caption{The architecture of the collision avoidance neural network. The network has the scan measurements $\mathbf{o}^t_z$, relative goal position $\mathbf{o}^t_g$ and current velocity $\mathbf{o}^t_v$ as inputs, and outputs the mean of velocity $\mathbf{v}^t_{mean}$. The final action $\mathbf a^t$ is sampled from the Gaussian distribution constructed by $\mathbf{v}^t_{mean}$  with a separated log standard deviation vector $\mathbf{v}^t_{logstd}$.} 
\label{fig:model} 
\vspace*{-0.2in}
\end{figure}

We design a 4-hidden-layer neural network as a non-linear function approximator to the policy $\pi_{\theta}$. Its architecture is shown in Figure~\ref{fig:model}. 
We employ the first three hidden layers to process the laser measurements $\mathbf{o}_z^t$ effectively. The first hidden layer convolves 32 one-dimensional filters with kernel size = 5, stride = 2 over the three input scans and applies ReLU nonlinearities~\cite{nair2010rectified}. The second hidden layer convolves 32 one-dimensional filters with kernel size = 3, stride = 2, again followed by ReLU nonlinearities. The third hidden layer is a fully-connected layer with 256 rectifier units. The output of the third layer is concatenated with the other two inputs ($\mathbf{o}_g^t$ and $\mathbf{o}_v^t$), and then are fed into the last hidden layer, a fully-connected layer with 128 rectifier units. The output layer is a fully-connected layer with two different activations: a sigmoid function is used to constrained the mean of translational velocity $v^t$ in $(0.0, 1.0)$ and the mean of rotational velocity $w^t$ in $(\minus 1.0, 1.0)$ through a hyperbolic tangent function ($\tanh$).   

Overall, the neural network maps the input observation vector $\mathbf{o}^t$ to a vector $\mathbf{v}_{mean}^t$. 
The final action $\mathbf{a}^t$ is sampled from a Gaussian distribution $\mathcal{N}(\mathbf{v}^t_{mean}, \mathbf{v}_{logstd}^t)$,
where  $\mathbf{v}^t_{mean}$ serves as the mean and $\mathbf{v}_{logstd}^t$ refers to a log standard deviation which will be updated solely during training.

\subsection{Multi-scenario multi-stage training}
\subsubsection{\textbf{Training algorithm}} 
\label{sec:train}
Even if deep reinforcement learning algorithms have been successfully applied in mobile robot motion planning, they have mainly focused on a discrete action space~\cite{zhu2017target,zhang2016deep} or on small-scale problems~\cite{tai2017virtual,chen2017decentralized,chen2017socially,kahn2017uncertainty}. 
Here we focus on learning a collision avoidance policy which performs robustly and effectively with a large number of robots in complex scenarios  with obstacles, such as corridors and mazes. We extend a recently proposed robust policy gradient algorithm, Proximal Policy Optimization (PPO)~\cite{schulman2017proximal,heess2017emergence,trpo}, to our multi-robot system. Our approach adapts the \textit{centralized learning, decentralized execution} paradigm. In particular, each robot receives its own $\mathbf{o}^t_i$ at each time step and executes the action generated from the shared policy $\pi_{\theta}$; the policy is trained with experiences collected by all robots simultaneously. 

As summarized in Algorithm~\ref{alg:ppo} (adapted from~\cite{schulman2017proximal,heess2017emergence}), the training process alternates between sampling trajectories by executing the policy in parallel, and updating the policy with the sampled data. During data collection, each robot exploits the same policy to generate trajectories until they collect a batch of data above $T_{max}$. Then the sampled trajectories are used to construct the surrogate loss $L^{PPO}(\theta)$, and this loss is optimized with the Adam optimizer~\cite{kingma2014adam} for $E_{\pi}$ epochs under the Kullback-Leiber (KL) divergence constraint. The state-value function $V_{\phi}(s_i^t)$, used as an baseline to estimate the advantage $\hat{A}_i^t$, is also approximated with a neural network with parameters $\phi$ on sampled trajectories. The network structure of $V_{\phi}$ is the same as that of the policy network $\pi_{\theta}$, except that it has only one unit in its last layer with a linear activation. We construct the squared-error loss $L^V(\phi)$ for $V_{\phi}$, and optimize it also with the Adam optimizer for $E_{V}$ epochs.
We update $\pi_{\theta}$ and $V_{\phi}$ independently and their parameters are not shared since we have found that using two separated networks will lead to better results in practice. 

This parallel PPO algorithm can be easily scaled to a large-scale multi-robot system with hundred robots in a decentralized fashion since each robot in the team is an independent worker collecting data. The decentralized execution not only dramatically reduce the time of sample collection, also make the algorithm suitable for training many robots in various scenarios. 
\begin{algorithm}
\caption{PPO with Multiple Robots}
\label{alg:ppo}
\begin{algorithmic}[1]
\State Initialize policy network $\pi_{\theta}$ and value function $V_{\phi}(s_t)$, and set hyperparameters as shown in Table~\ref{tab:parameters}. 
\For {$\text{iteration} = 1, 2,..., $}
  \State // \textit{Collect data in parallel}
  \For{$\text{robot } i = 1, 2, ... N$}
  \State Run policy $\pi_{\theta}$ for $T_i$ timesteps, collecting $\{ \mathbf{o}_i^t, r_i^t, \mathbf{a}_i^t \} $, where $t \in [0, T_i]$
  \State Estimate advantages using GAE~\cite{schulman2015high} $\hat{A}_i^t = {\sum}_{l=0}^{T_i} ({\gamma \lambda})^l {\delta}_{i}^{t} $, where ${\delta}_{i}^{t} = r_i^t + \gamma V_{\phi}(s_i^{t+1}) - V_{\phi}(s_i^{t})$
   \State {\bf break}, if $\sum _{i=1}^N T_i > T_{max}$
  \EndFor
  \State $\pi_{old} \gets \pi_{\theta} $ 
  \State // \textit{Update policy}
  \For{$j = 1,..., E_{\pi}$}
    \State $ L^{PPO}(\theta) = \sum _{t=1}^{T_{max}} \frac{\pi_{\theta}(a_i^t \mid o_i^t)}{\pi_{old}(a_i^t \mid o_i^t)} \hat{A}_i^t - \beta \mathrm{KL}[\pi_{old} \mid \pi_{\theta}]+ \xi \mathrm{max}(0, \mathrm{KL}[\pi_{old} \mid \pi_{\theta}] - 2\mathrm{KL}_{target})^2 $ 
    \If{$\mathrm{KL}[\pi_{old} \mid \pi_{\theta}] > 4\mathrm{KL}_{target}$}
    	\State {\bf break} and continue with next iteration $i+1$
    \EndIf{}
    \State Update $\theta$ with $lr_{\theta}$ by Adam~\cite{kingma2014adam} w.r.t $L^{PPO}(\theta)$ 
  \EndFor
  \State // \textit{Update value function}
  \For{$k = 1,..., E_{V}$}
    \State $ L^{V}(\phi) = -\sum _{i=1}^N \sum _{t=1}^{T_{i}}(\sum _{t'>t}\gamma ^{t'-t}r_i^{t'} - V_{\phi}(s_i^t))^2$
    \State Update $\phi$ with $lr_{\phi}$ by Adam w.r.t $L^{V}(\phi)$ 
  \EndFor
  \State // \textit{Adapt KL Penalty Coefficient}
  \If{$\mathrm{KL}[\pi_{old} \mid \pi_{\theta}] > \beta _{high} \mathrm{KL}_{target}$}
  	\State $\beta \gets \alpha \beta$
  \ElsIf{$\mathrm{KL}[\pi_{old} \mid \pi_{\theta}] < \beta _{low} \mathrm{KL}_{target}$}
  	\State $\beta \gets \beta / \alpha$
  \EndIf
\EndFor
\end{algorithmic}
\end{algorithm}

\subsubsection{\textbf{Training scenarios}} 
\begin{figure}[t] 
\centering
\includegraphics[width=1\linewidth]{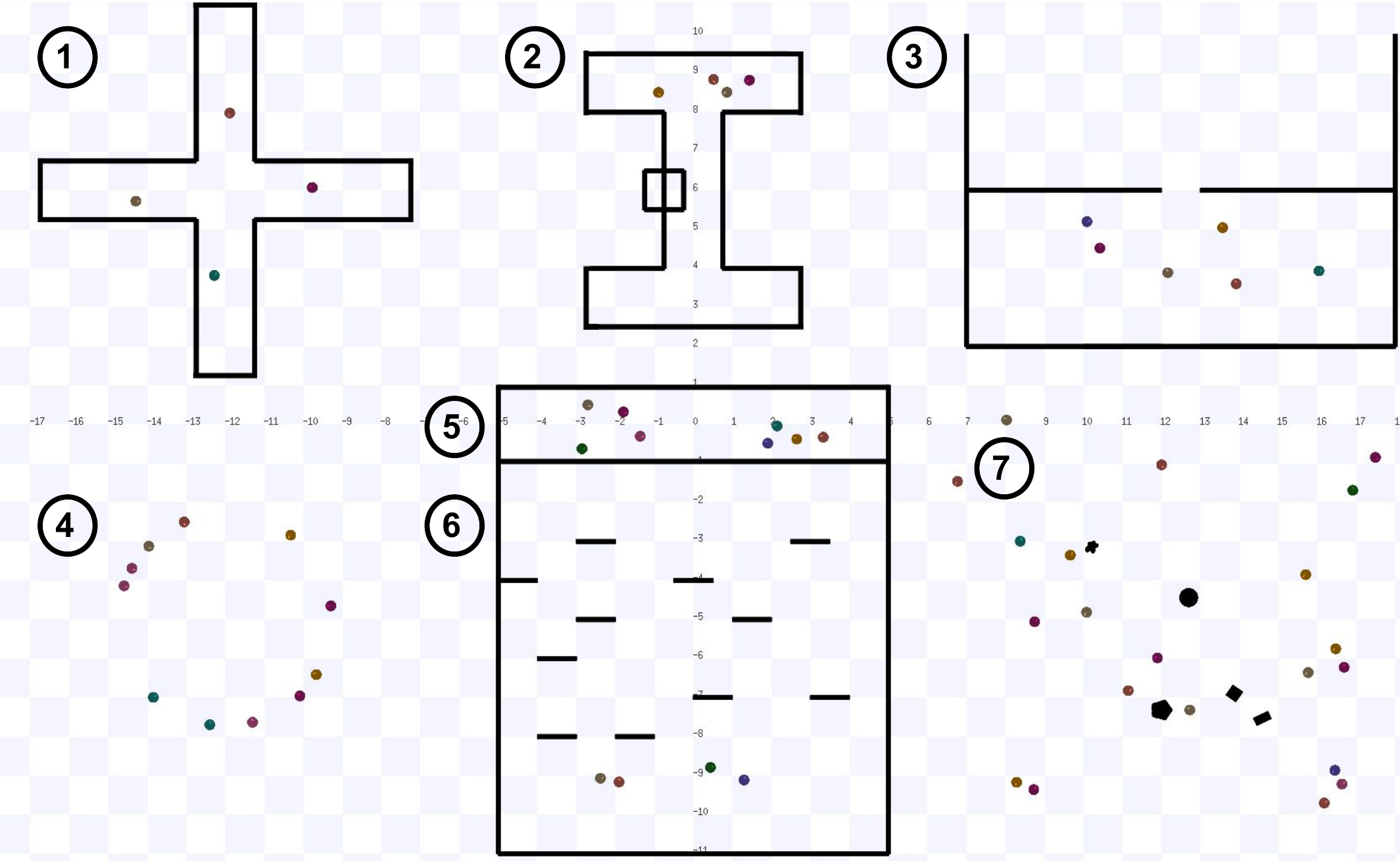}
\caption{Scenarios used to train the collision avoidance policy. All robots are modeled as a disc with the same radius. Obstacles are shown in black.}
\label{fig:scene}
\end{figure}

To expose our robots to diverse environments, we create different scenarios with a variety of obstacles using the Stage mobile robot simulator\footnote{\url{http://rtv.github.io/Stage/}} (as shown in Figure~\ref{fig:scene}) and move all robots concurrently. In scenario 1, 2, 3, 5, and 6 in Figure~\ref{fig:scene} (black solid lines are obstacles), we first select reasonable starting and arrival areas from the available workspace, then randomly sample the start and goal positions for each robot in the corresponding areas. Robots in scenario 4 are randomly initialized in a circle with a varied radius, and they aim to reach their antipodal positions by crossing the central area.  As for scenario 7, we generate random  positions for both robots and obstacles (shown in black) at the beginning of each episode; and the target positions of robots are also randomly selected. These rich, complex training scenarios enable robots to explore their high-dimensional observation space and are likely to improve the quality and robustness of the learned policy. Combining with the \textit{centralized learning, decentralized execution} mechanism, the collision avoidance policy is effectively optimized at each iteration over a variety of environments. 

\subsubsection{\textbf{Training stages}} 
\begin{figure} 
\centering
\includegraphics[width=1\linewidth]{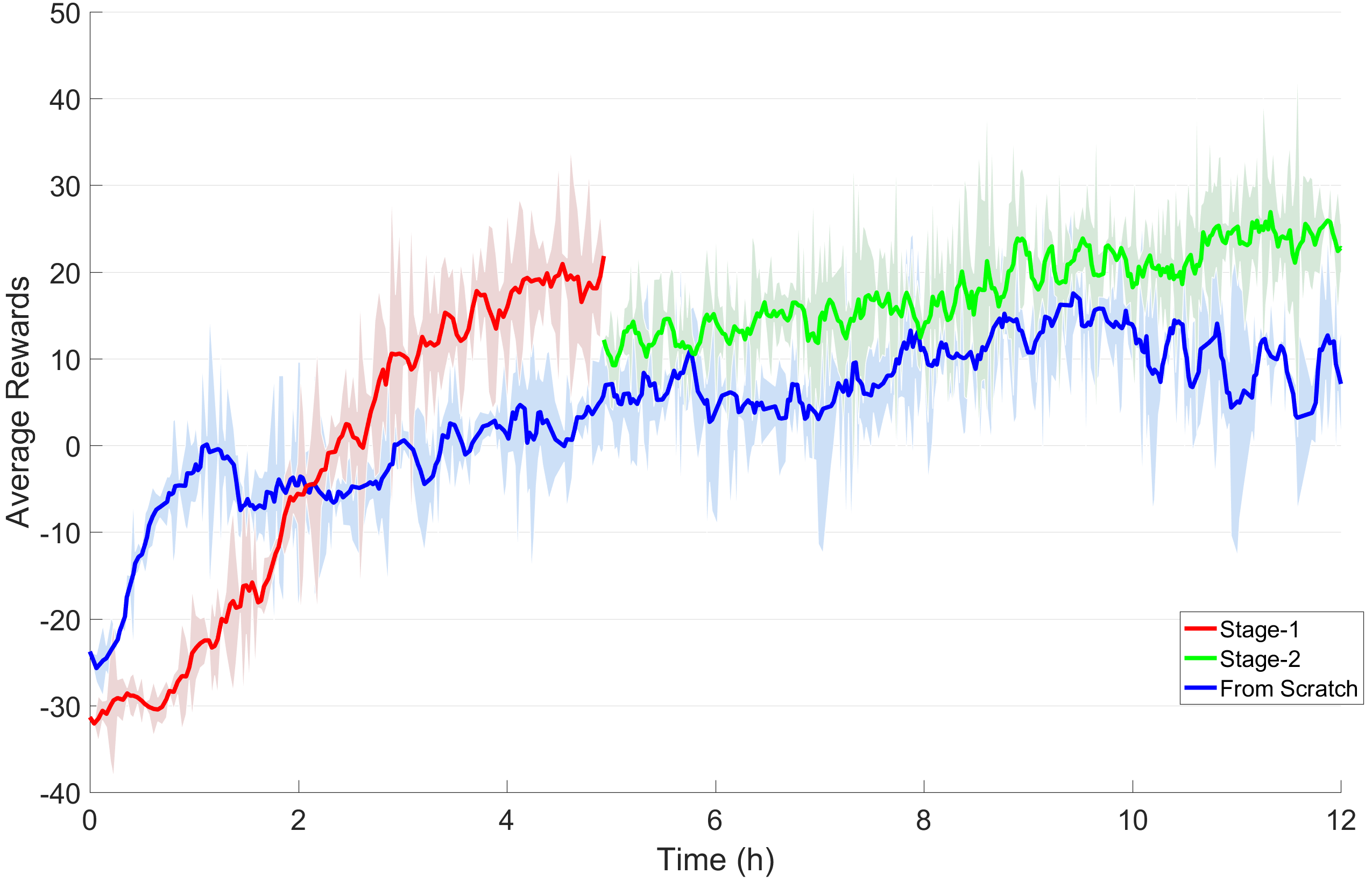}
\caption{Average rewards shown in wall time during the training process.}
\label{fig:reward}
\vspace*{-0.3in}
\end{figure}

Although training on multiple environments simultaneously brings robust performance over different test cases (see Section~\ref{sec:gen}), it makes the training process harder. Inspired by the curriculum learning paradigm~\cite{bengio2009curriculum}, we propose a two-stage training process, which accelerates the policy to converge to a satisfying solution, and gets higher rewards than the policy trained from scratch with the same number of epoch (as shown in Figure~\ref{fig:reward}). In the first stage, we only train 20 robots on the random scenarios (scenario 7 in Figure~\ref{fig:scene}) without any obstacles, this allows our robots learn fast on relatively simple collision avoidance tasks. Once the robots achieve reliable performance, we stop the Stage 1 and save the trained policy. The policy will continue to be updated in the Stage 2, where the number of robots increases to 58 and they are trained on richer and more complex scenarios shown in Figure~\ref{fig:scene}. 

\section{Experiments and Results}
\label{sec:exp}
In this section, we first describe the hyper-parameters and computational complexity of the training process. Then, we quantitatively compare our policy with other methods in various simulated scenarios. Lastly, we demonstrate the good generalization capability of the learned policy in several challenging and complex environments. 

\subsection{Training configuration and computational complexity}
The implementation of our algorithm is in TensorFlow and the large-scale robot group with laser scanners is simulated in the Stage simulator. We train the multi-robot collision avoidance policy on a computer with an i7-7700 CPU and a Nvidia GTX 1080 GPU. The offline training takes 12 hours (about 600 iterations in Algorithm~\ref{alg:ppo}) for the policy to converge to a robust performance in all scenarios. The hyper-parameters in Algorithm~\ref{alg:ppo} is summarized in Table~\ref{tab:parameters}. In particular, the learning rate $lr_{\theta}$ of policy network is set to 5$e$\minus5 in the first stage, and is then reduced to 2$e$\minus5 in the second training stage. As for running ten robots in the online decentralized control, it takes 3ms for the policy network to compute new actions on the CPU and about 1.3ms on the GPU. 

\begin{table}
 \begin{tabularx}{0.5\textwidth}{l|X}
   \hline
  Parameter & Value  \\
   \hline
   \text{$\lambda$ in line 6} & 0.95  \\
  \text{$\gamma$ in line 6 and 20} & 0.99 \\
   \text{$T_{max}$ in line 7} & 8000 \\
   \text{$E_{\phi}$ in line 11} & 20 \\
   \text{$\beta$ in line 12} & 1.0 \\
   \text{$\mathrm{KL}_{target}$ in line 12} & $15e\minus 4$ \\
   \text{$\xi$ in line 12} & 50.0 \\
   \text{$lr_{\theta}$ in line 16} & $5e\minus 5$ (first stage), $2e\minus 5$ (second stage) \\
   \text{$E_V$ in line 19} & 10 \\
   \text{$lr_{\phi}$ in line 21} & $1e\minus 3$ \\ 
   \text{$\beta_{high}$ in line 24} & 2.0 \\
   \text{$\alpha$ in line 24 and 27} & 1.5 \\  
   \text{$\beta_{low}$ in line 26} & 0.5 \\
   \hline
 \end{tabularx}
\caption{The hyper-parameters of our training algorithm described in Algorithm~\ref{alg:ppo}.}
\label{tab:parameters}
\vspace*{-0.2in}
\end{table}

\subsection{Quantitative comparison on various scenarios}
\subsubsection{\textbf{Performance metrics}}
To compare the performance of our policy with other methods over various test cases, we use the following performance metrics. For each method, every test case is evaluated for $50$ repeats. 
\renewcommand{\labelitemi}{\textbullet}
\begin{itemize}
\item \textit{Success rate} is the ratio of the number of robots reaching their goals within a certain time limit without any collisions over the total number of robots. 
\item \textit{Extra time $\bar{t}_e$} measures the difference between the travel time averaged over all robots and the lower bound of the travel time (i.e. the average cost time of going straight toward the goal for robots at max speed~\cite{godoy2016implicit,chen2017decentralized}).
\item \textit{Extra distance $\bar{d}_e$} measures the difference between the average traveled trajectory length of the robots and the lower bound of traveled distance of the robots (i.e. the average traveled distance for robots following the shortest paths toward the goals).
\item \textit{Average speed $\bar{v}$} measures the average speed of the robot team during  navigation.
\end{itemize}
Note that the extra time $\bar{t}_e$ and extra distance $\bar{d}_e$ metrics are measured over all robots during the evaluation which remove the effects due to the variance in the number of agents and the different distance to goals.

\subsubsection{\textbf{Circle scenarios}}
We first compare our multi-scenario multi-stage learned policy with the NH-ORCA policy~\cite{alonso2013optimal}, and the policy trained using supervised learning (SL-policy, a variation of ~\cite{long2017deep}, see below for details) on circle scenarios with different number of robots. The \textit{circle} scenarios are similar to the scenario 4 shown in Figure~\ref{fig:scene}, except we set robots uniformly on the circle. We use the open-sourced NH-ORCA implementation
from~\cite{hennes2012multi,claes2012collision}, and share the ground truth positions and velocities for all robots in the simulations. The policy learned in supervised mode, with the same architecture of our policy (described in Section~\ref{sec:model}), is trained on about 800,000 samples with the method adapted from~\cite{long2017deep,pfeiffer2017perception}.   

Compared to the NH-ORCA policy, our learned policy has significant improvement over it in terms of success rate, average extra time and travel speed. Although our learned policy has a slightly longer traveled paths than the NH-ORCA policy in scenarios with robot number above 15 (the third row in Table~\ref{tab:1circle}), the larger speed (the fourth row in Table~\ref{tab:1circle}) helps our robots reach their goals more quickly. Actually the slightly longer path is a byproduct of the higher speed since the robots need more space to decelerate before stopping at goals. 
\begin{table*}
  \resizebox{\textwidth}{!}{  
 \begin{tabularx}{1\textwidth}{l|l|l|l|l|l|l|l|l}
    \hline
  	Metrics & Method & 4 & 6 & 8 & 10 & 12 & 15 & 20 \\
    \hline
   	\multirow{3}{*}{Success Rate} & SL-policy & 0.6  & 0.7167 & 0.6125 & 0.71 & 0.6333 & - & -  \\
      & NH-ORCA & \textbf{1.0}  & 0.9667 & 0.9250 & 0.8900 & 0.9000 & 0.8067 & 0.7800 \\
      & Our policy & \textbf{1.0}  & \textbf{1.0} & \textbf{1.0} & \textbf{1.0} & \textbf{1.0} & \textbf{1.0} & \textbf{0.9650} \\
    \hline
    \multirow{3}{*}{Extra Time} & SL-policy & 9.254 / 2.592  & 9.566 / 3.559 & 12.085 / 2.007 & 13.588 / 1.206 & 19.157 / 2.657 & - & -  \\
      & NH-ORCA & 0.622 / 0.080  & 0.773 / 0.207 & 1.067 / 0.215 & 0.877 /0.434 & 0.771 / 0.606 & 1.750 / 0.654 & 1.800 / 0.647 \\
      & Our policy & \textbf{0.148 / 0.004}  & \textbf{0.193 / 0.006} & \textbf{0.227 / 0.005} & \textbf{0.211 / 0.007} & \textbf{0.271 / 0.005} & \textbf{0.350 / 0.014} & \textbf{0.506 / 0.016} \\
      \hline
      \multirow{3}{*}{Extra Distance} & SL-policy & 0.358 / 0.205  & 0.181 / 0.146 & 0.138 / 0.079 & 0.127 / 0.047 & 0.141 / 0.027 & - & -  \\
      & NH-ORCA & \textbf{0.017 / 0.004}  & \textbf{0.025 / 0.005} & 0.041 / 0.034 & \textbf{0.034 / 0.009} & 0.062 / 0.024 & \textbf{0.049 / 0.019} & \textbf{0.056 / 0.018} \\
      & Our policy & \textbf{0.017 / 0.007}  & 0.045 / 0.002 & \textbf{0.040 / 0.001} & 0.051 / 0.001 & \textbf{0.056 / 0.002} & 0.062 / 0.003 & 0.095 / 0.007 \\
      \hline
      \multirow{3}{*}{Average Speed} & SL-policy & 0.326 / 0.072 & 0.381 / 0.087 & 0.354 / 0.042 & 0.355 / 0.022 & 0.308 / 0.028 & - & -  \\
      & NH-ORCA & 0.859 / 0.012  & 0.867 / 0.026 & 0.839 / 0.032 & 0.876 / 0.045 & 0.875 / 0.054 & 0.820 / 0.052 & 0.831 / 0.042 \\
      & Our policy & \textbf{0.956 / 0.006}  & \textbf{0.929 / 0.002} & \textbf{0.932 / 0.001} & \textbf{0.928 / 0.002} & \textbf{0.921 / 0.002} & \textbf{0.912 / 0.003} & \textbf{0.880 / 0.006} \\
      \hline
 \end{tabularx}
 }
\caption{Performance metrics (as mean/std) evaluated for different methods on the circle scenarios with different number of robots.}
\label{tab:1circle}
\vspace*{-0.2in}
\end{table*}





\subsubsection{\textbf{Random scenarios}}
\begin{figure}[h] 
\centering
\begin{subfigure}{0.22\textwidth}
\includegraphics[width=3.5cm,height=3cm]{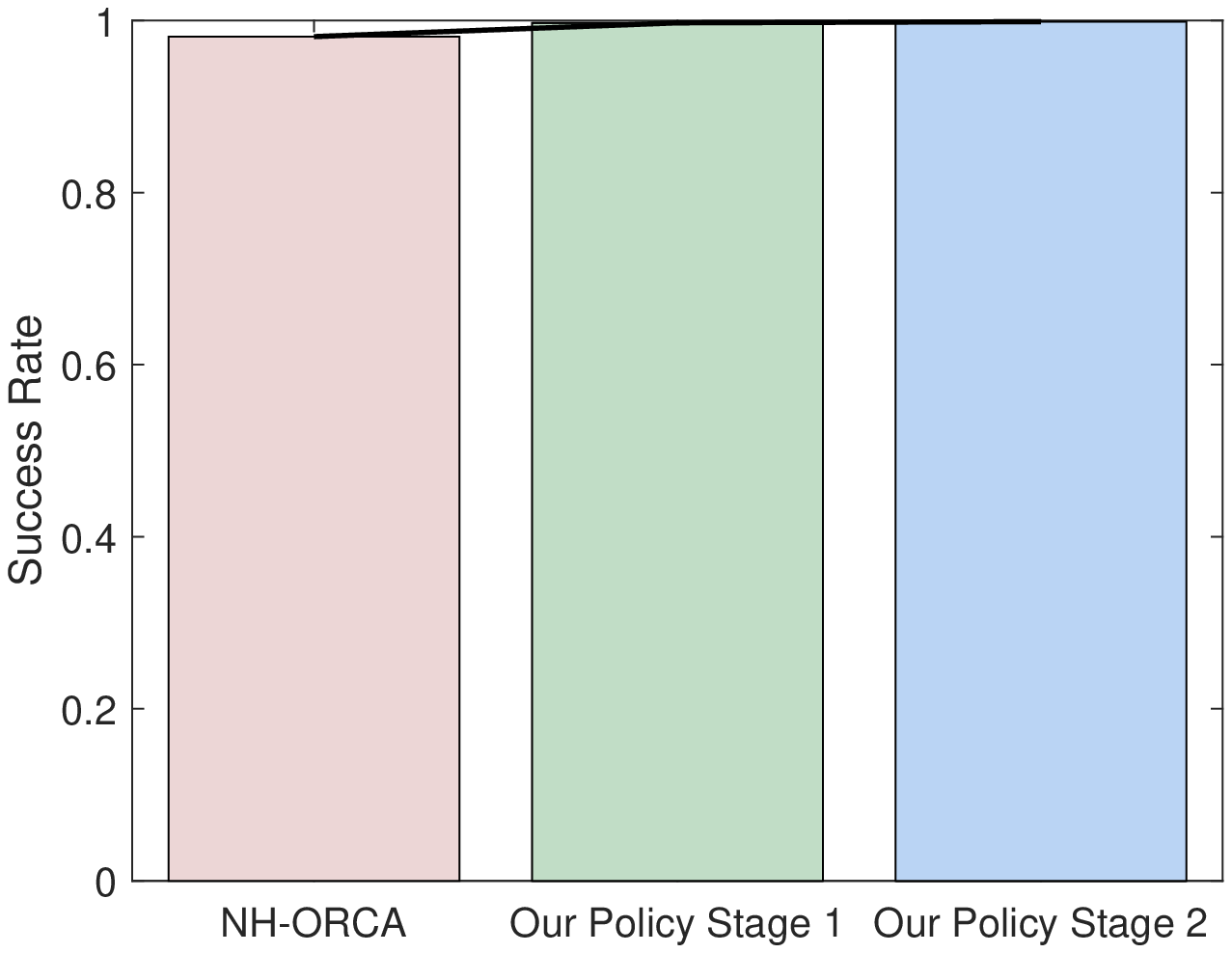}
\caption{Success rate}
\label{fig:2succ}
\end{subfigure}
\begin{subfigure}{0.22\textwidth}
\includegraphics[width=3.5cm,height=3cm]{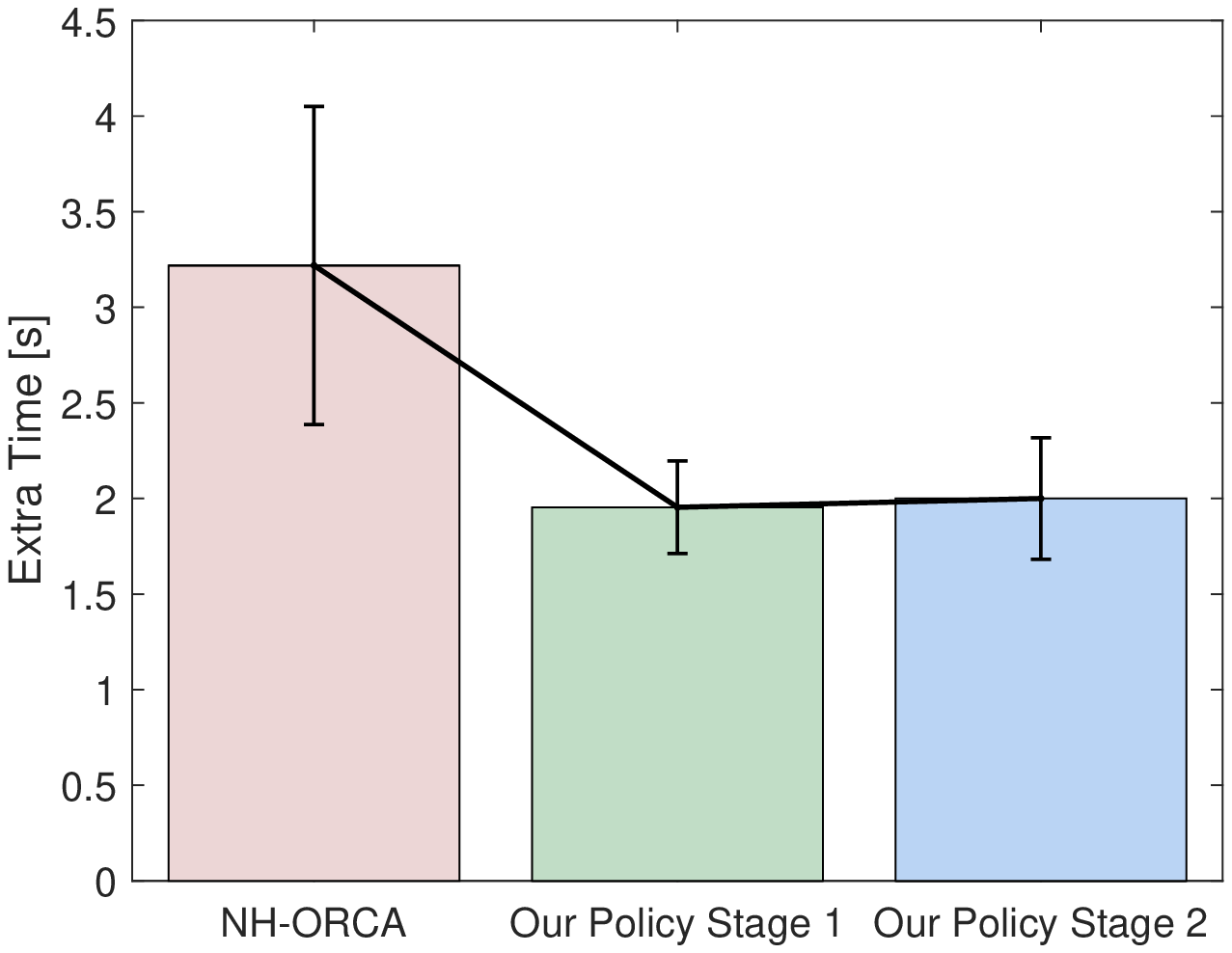}
\caption{Extra time}
\label{fig:2time}
\end{subfigure} \\
\centering
\begin{subfigure}{0.22\textwidth}
\includegraphics[width=3.5cm,height=3cm]{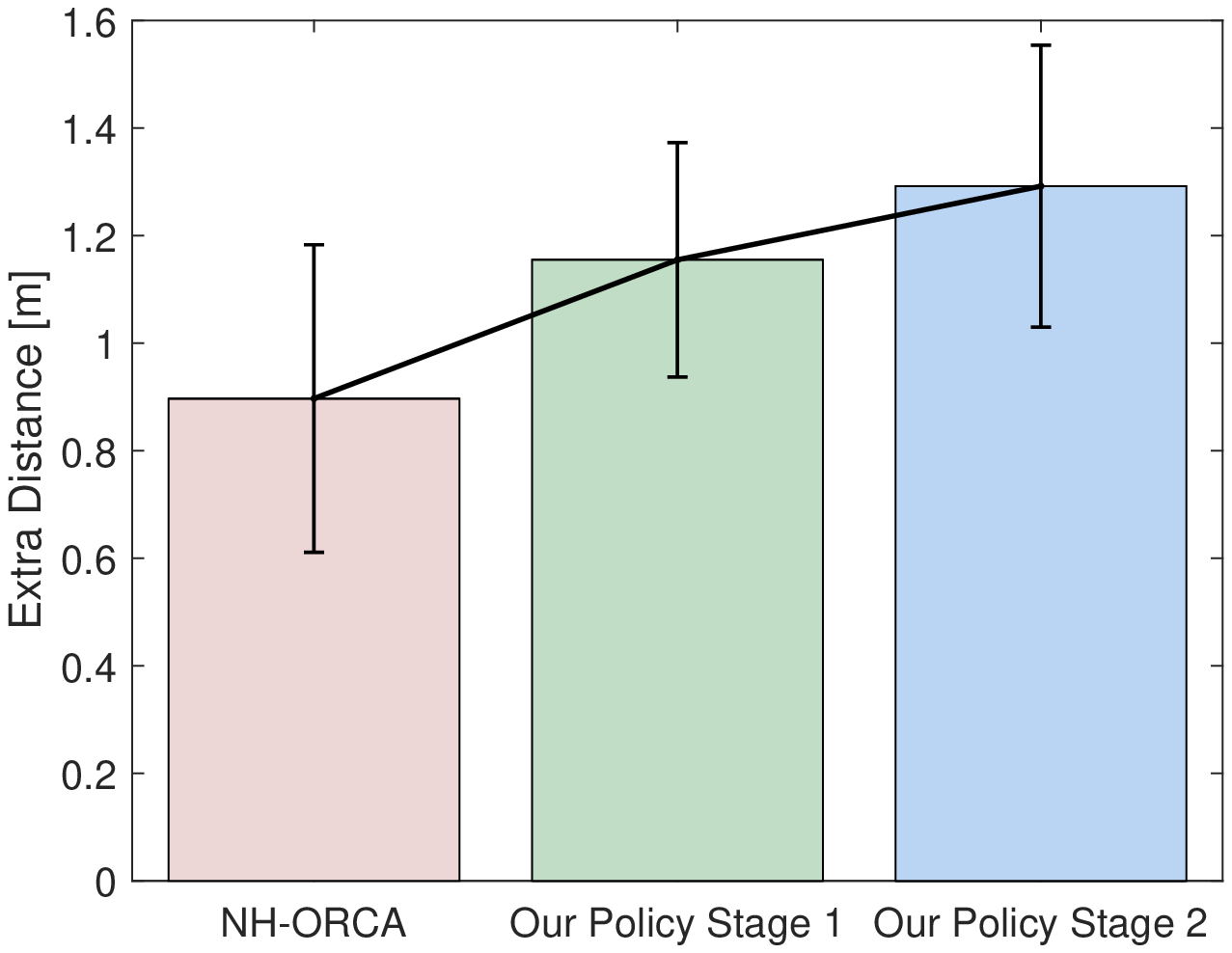}
\caption{Extra distance}
\label{fig:2dist}
\end{subfigure}
\begin{subfigure}{0.22\textwidth}
\includegraphics[width=3.5cm,height=3cm]{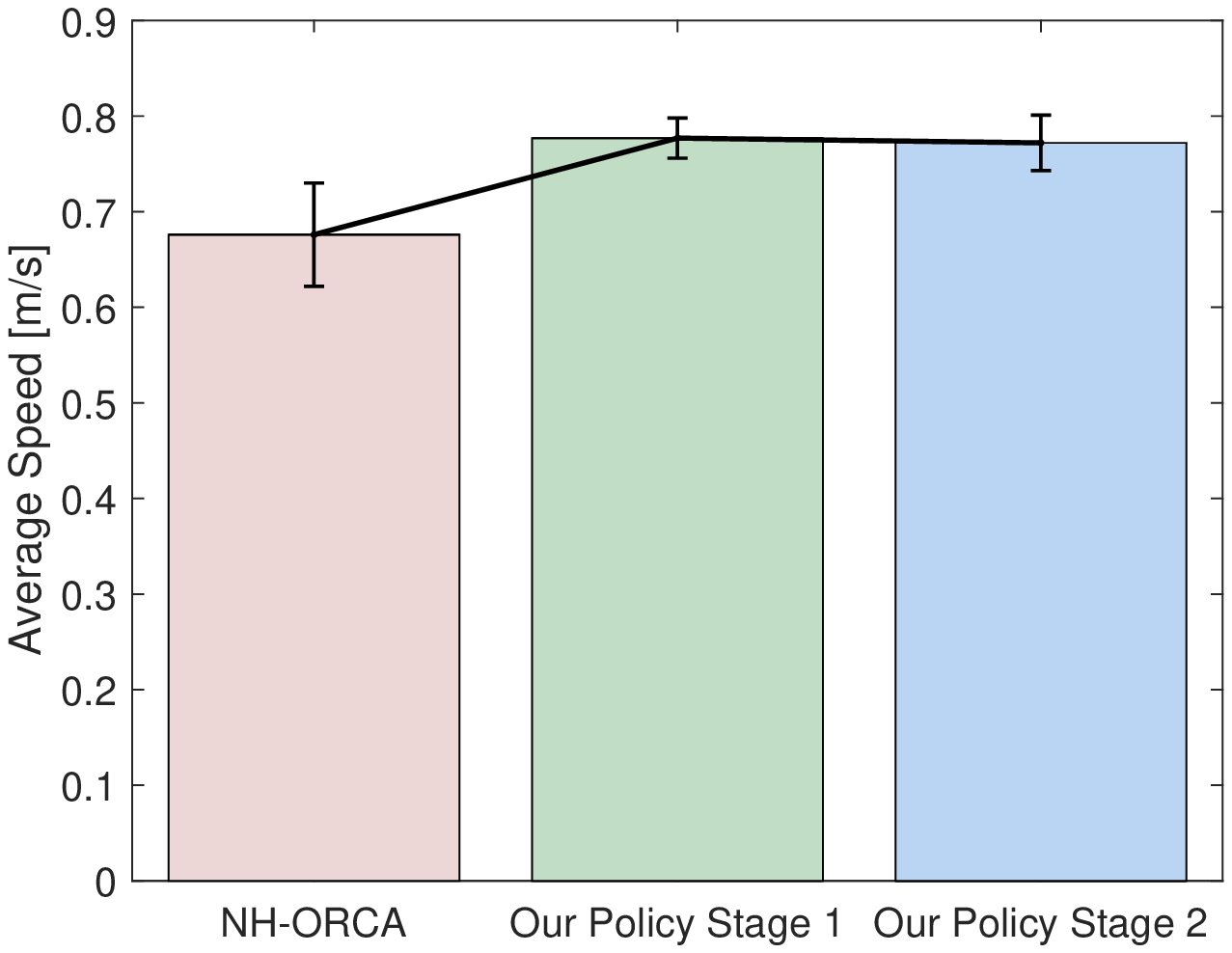}
\caption{Average speed}
\label{fig:2speed}
\end{subfigure}
\caption{Performance metrics evaluated for our learned policies and the NH-ORCA policy on random scenarios. }
\label{fig:2random}
\vspace*{-0.1in}
\end{figure}

Random scenarios are other frequently used scenes to evaluate the performance of multi-robot collision avoidance. To measure the performance of our method on random scenarios (as shown in the 7th scenario in Figure~\ref{fig:scene}), we first create 5 different random scenarios with 15 robots in each case. For each random scenario, we repeat our evaluations 50 times. The results are shown in Figure~\ref{fig:2random}, which compares our final policy with the policy only trained in the stage 1 (Section~\ref{sec:train}) and the NH-ORCA policy. We can observe that both policies trained using deep reinforcement learning have higher success rate than NH-ORCA policy (Figure~\ref{fig:2succ}). It can also be seen that robots using the learned policies (in Stage 1 and 2) are able to reach their targets much faster than that of NH-ORCA (Figure~\ref{fig:2time}). Although the learned policies have longer trajectory length (Figure~\ref{fig:2dist}), the higher average speed (Figure~\ref{fig:2speed}) and success rate indicate that our policies enable a robot to better anticipate other robots' movements. Similar to the circle scenarios above, the slightly longer path is due to robots' requirement to decelerate before stopping at goals. In addition, the stage-1 policy's high performance in the random scenario is partially contributed by the overfitting since it is trained in a similar random scenario, while the stage-2 policy is trained in multiple scenarios.

\subsubsection{\textbf{Group scenarios}}
\begin{figure}[!h] 
\begin{subfigure}{0.24\textwidth}
\includegraphics[width=1.0\linewidth,height=1.2cm]{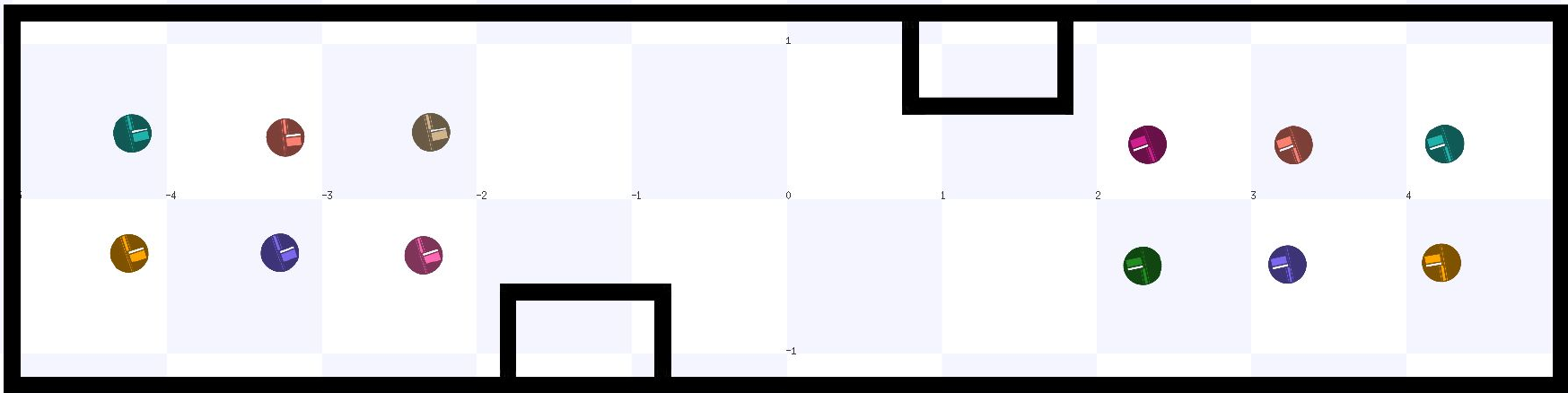}
\caption{Corridor scenario}
\label{fig:3corridor_scene}
\end{subfigure}
\begin{subfigure}{0.24\textwidth}
\includegraphics[width=1.0\linewidth,height=1.15cm]{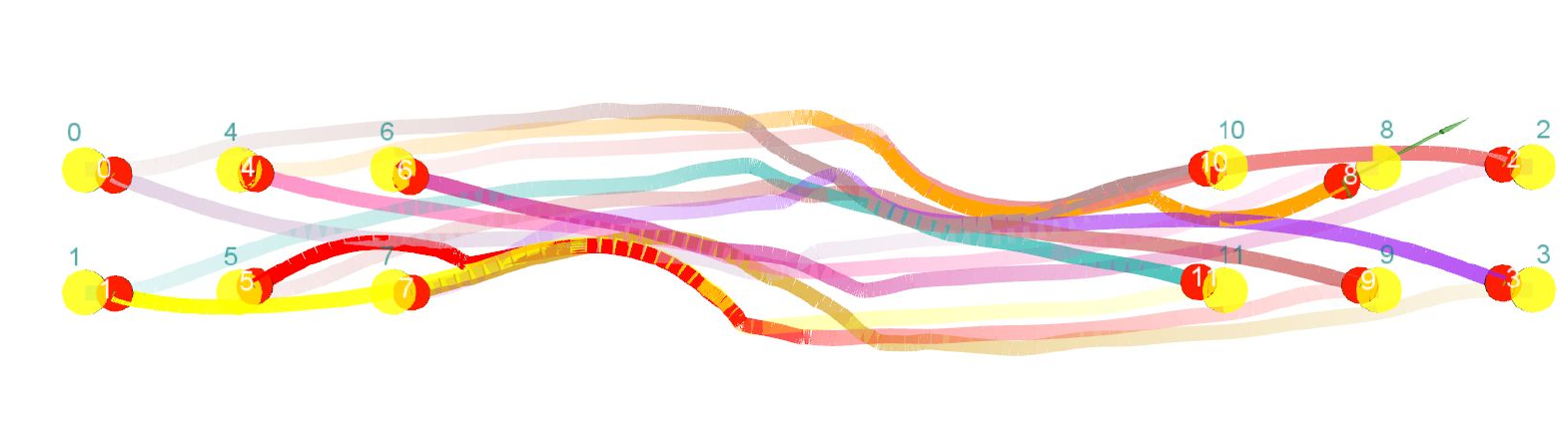}
\caption{Robot trajectories}
\label{fig:3corridor_path}
\end{subfigure}
\caption{Two group of robots moving in a corridor with obstacles. (a) shows the corridor scenario. (b) shows trajectories generated by our Stage-2 policy.}
\label{fig:3corridor}
\vspace*{-0.2in}
\end{figure}

\begin{figure}[!h]
\centering
\includegraphics[width=1\linewidth]{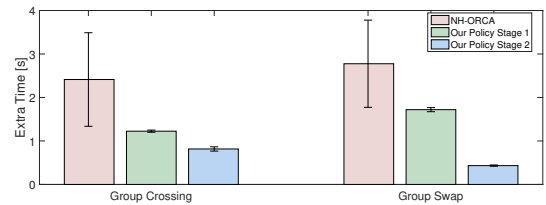}
\caption{Extra time $\bar{t}_e$ of our policies (Stage 1 and Stage 2) and the NH-ORCA policy on two group scenarios.}
\label{fig:3group}
\vspace*{-0.2in}
\end{figure}

In order to evaluate the cooperation between robots, we would like to test our trained policy on more challenging scenarios, e.g. group swap, group crossing and group moving in the corridors. In the group swap scenarios, we navigate two groups of robots (each group has 6 robots) moving in opposite directions to swap their positions. As for group crossing scenarios, robots are separated in two groups, and their paths will intersect in the center of the scenarios.  
We compare our method with NH-ORCA on these two cases by measuring the average extra time $\bar{t}_e$ with 50 trials. It can be seen from Figure~\ref{fig:3group} that our policies performs much better than NH-ORCA on both cases. The shorter goal-reached time demonstrates that our policies have learned to produce more cooperative behaviors than reaction-based methods (NH-ORCA). We then evaluate three policies on the corridor scene, where two groups exchange their positions in a narrow corridor with two obstacles as shown in Figure~\ref{fig:3corridor_scene}. However, only the Stage-2 policy can complete this challenging task (with paths as illustrated in Figure~\ref{fig:3corridor_path}).The failure of the Stage-1 policy shows that the co-training on a wide range of scenarios can lead to robust performance across different situations. 
NH-ORCA policy fails on this case because it relies on global planners to guide robots navigating in the complex environment. As mentioned in Section~\ref{sec:intro}, the agent-level collision avoidance policy (e.g. NH-ORCA) requires extra pipeline (e.g. a grid map indicating the obstacles) to explicitly identify and process the static obstacles while our method (sensor-level policy) implicitly infers the obstacles from raw sensor readings without any additional processing. 

\subsection{Generalization}
\label{sec:gen}
\begin{figure}[h] 
\begin{subfigure}{0.22\textwidth}
\includegraphics[width=1.0\linewidth]{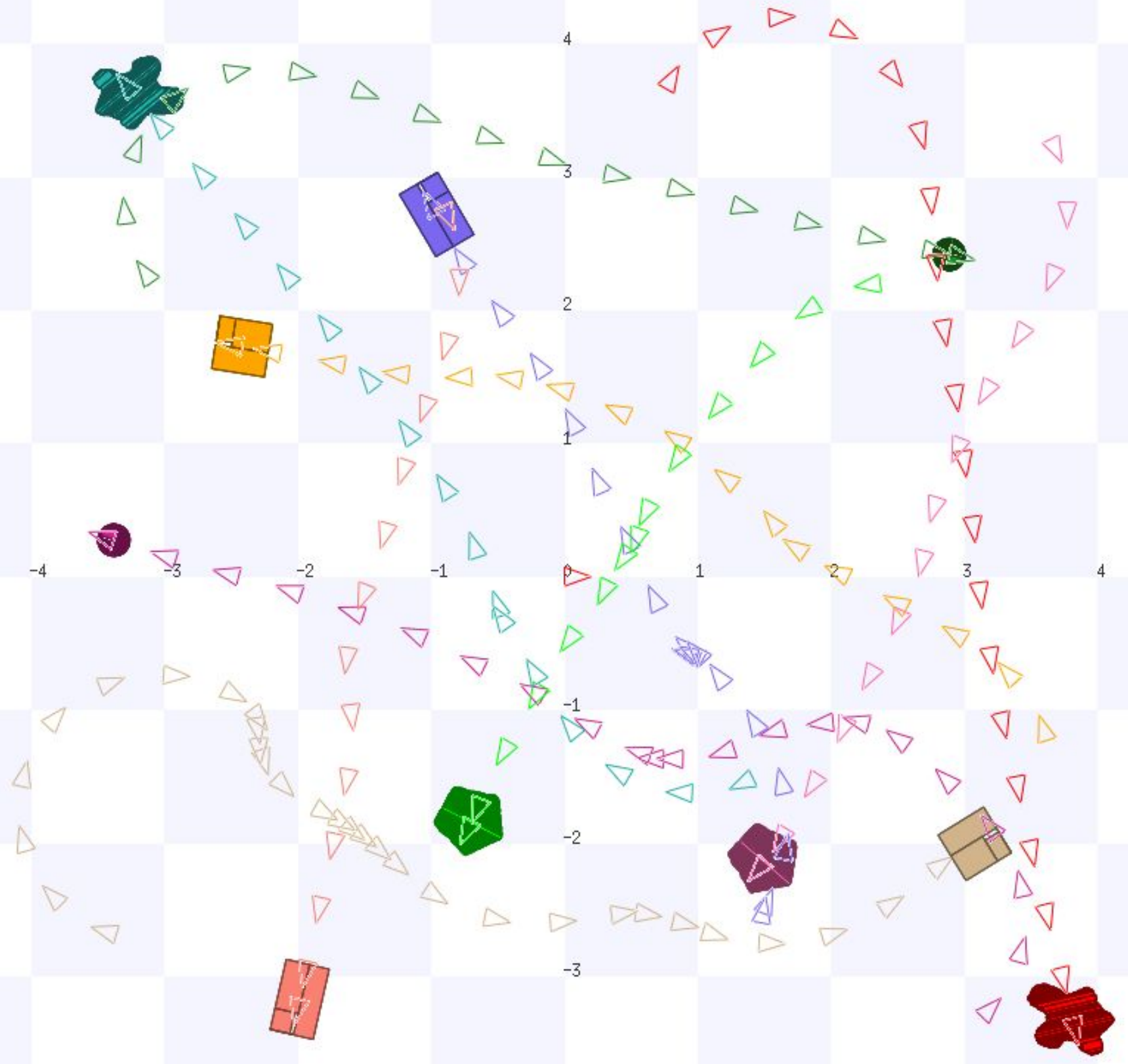}
\caption{Heterogeneous robots}
\label{fig:hete}
\end{subfigure}
\begin{subfigure}{0.22\textwidth}
\includegraphics[width=1.0\linewidth]{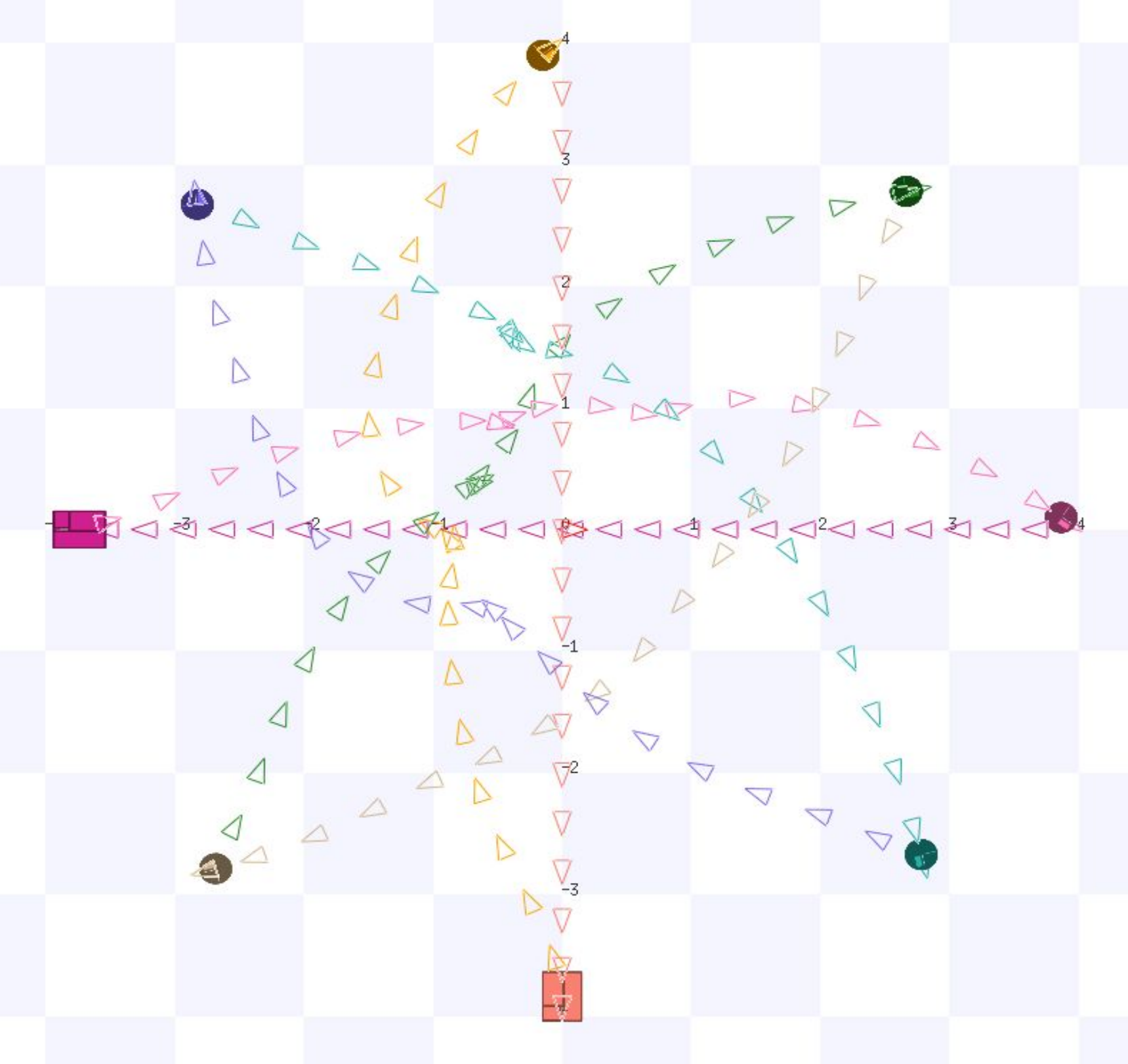}
\caption{Non-cooperative robots}
\label{fig:dynamic}
\end{subfigure}
\caption{In the heterogeneous robot team (a), only the two disc-shaped robots are used in training. (b) shows 6 robots moving around two non-cooperative robots (rectangle-shaped), which traveled in straight lines with a fast speed.}
\label{fig:gene}
\vspace*{-0.1in}
\end{figure}

\begin{figure}[h] 
\begin{subfigure}{0.22\textwidth}
\includegraphics[width=1.0\linewidth]{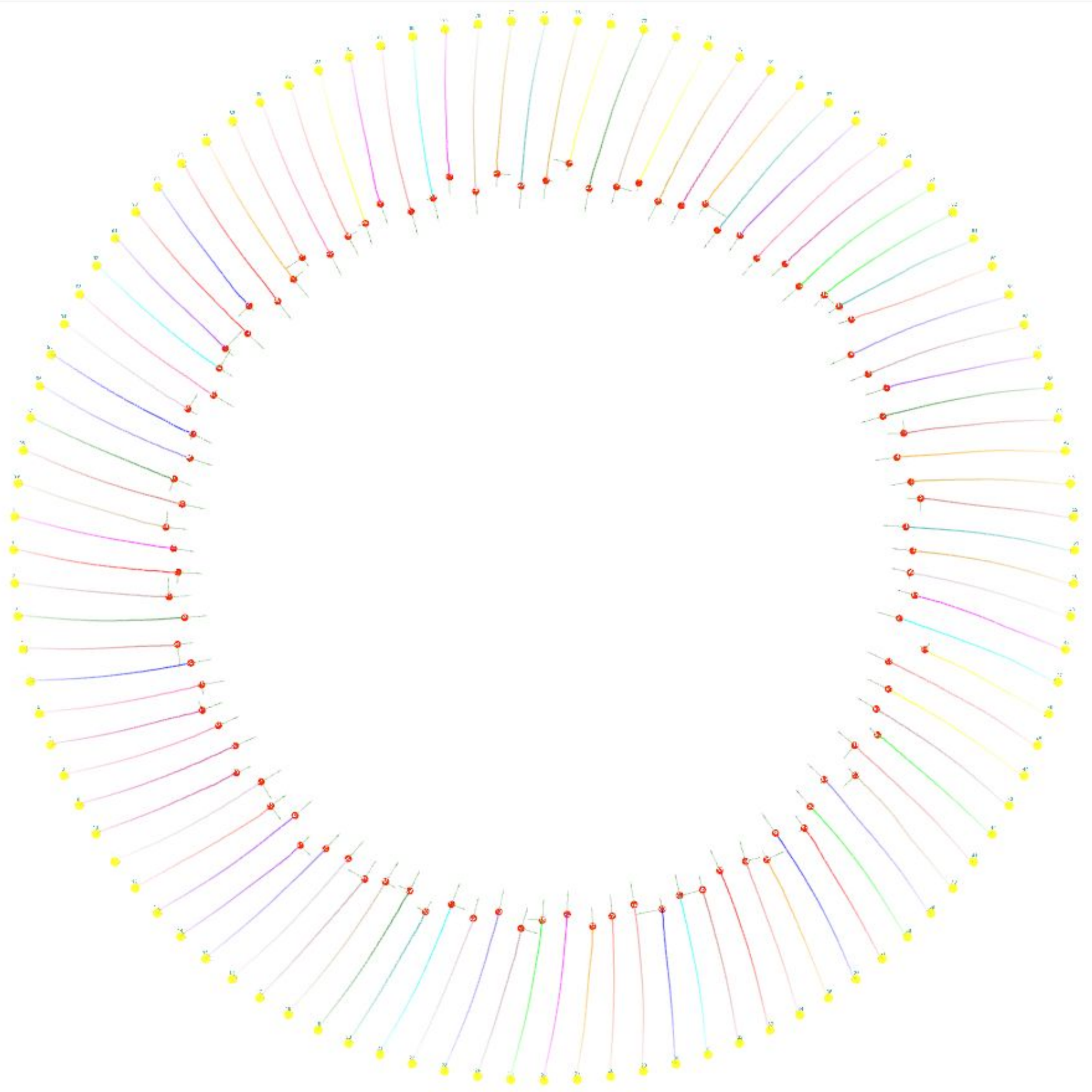}
\label{fig:100start}
\end{subfigure}
\begin{subfigure}{0.22\textwidth}
\includegraphics[width=1.0\linewidth]{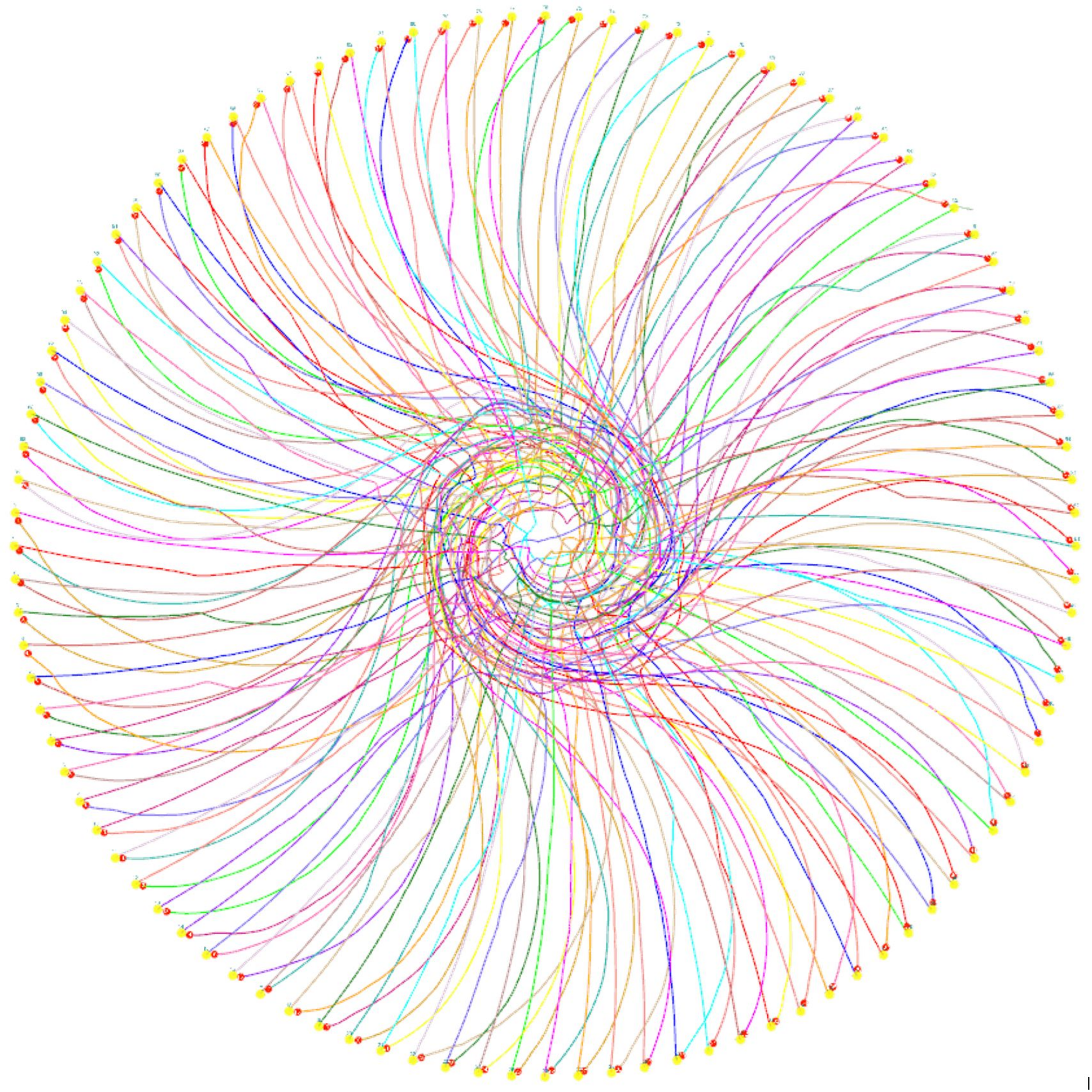}
\label{fig:100end}
\end{subfigure}
\caption{Simulation of 100 robots trying to move through the center of a circle to antipodal positions.}
\label{fig:100}
\vspace*{-0.1in}
\end{figure}

A notable feature benefited from multi-scenario training is the good generalization capability of the learned policy (Stage-2 policy). 
As mentioned in Section~\ref{sec:prob}, our policy is trained within a robot team where all robots share the same collision avoidance strategy. Non-cooperative robots are not introduced over the entire training process. Interestingly, the result shown in Figure~\ref{fig:dynamic} demonstrates that the learned policy can directly generalize well to avoid non-cooperative agents (i.e. the rectangle-shaped robots in Figure~\ref{fig:dynamic} which travel in straight lines with a fixed speed). 
Recall our policy is trained on robots with the same shape and a fixed radius. Figure~\ref{fig:hete} exhibits that the learned policy can also navigate efficiently a heterogeneous group of robots consisting of robots with different sizes and shapes to reach their goals without any collisions. 
To test the performance of our method on large-scale scenarios, we simulate 100 robots in a large circle moving to antipodal positions as shown in Figure~\ref{fig:100}. It shows that our learned policy can directly generalize to large-scale environments without any fine-tuning.

\section{Conclusion}
\label{sec:conclusion}

In this paper, we present a multi-scenario multi-stage training framework to optimize a fully decentralized sensor-level collision avoidance policy with a robust policy gradient algorithm. The learned policy has demonstrated several advantages on an extensive evaluation of the state-of-the-art NH-ORCA policy in terms of success rate, collision avoidance performance, and generalization capability. 
Our work can serve as a first step towards reducing the navigation performance gap between the centralized and decentralized methods, though we are fully aware that the learned policy focusing on local collision avoidance cannot replace a global path planner when scheduling many robots to navigate through complex environments with dense obstacles.

{\small
\bibliographystyle{IEEEtran}
\bibliography{references}
}

\end{document}